\documentclass[10pt,twocolumn,letterpaper]{article}

\usepackage[pagenumbers]{cvpr}      
\definecolor{cvprblue}{rgb}{0.21,0.49,0.74}
\usepackage[pagebackref,breaklinks,colorlinks,allcolors=cvprblue]{hyperref}
\usepackage{algorithm}
\usepackage{algorithmic}
\usepackage{xcolor}
\usepackage{graphicx}
\usepackage{amsmath}
\usepackage{capt-of}
\usepackage{adjustbox}
\usepackage{cuted}

\usepackage{xcolor} 
\usepackage{colortbl} 
\usepackage{graphicx} 
\usepackage{booktabs}
\usepackage{color}
\usepackage[utf8]{inputenc}
\usepackage{array}
\usepackage{multirow}
\usepackage{amssymb} 
\usepackage{tabularx}
\usepackage{tikz}

\newcommand{\figref}[1]{Fig.~\ref{#1}}
\newcommand{\tabref}[1]{Tab.~\ref{#1}}
\newcommand{\secref}[1]{Sec.~\ref{#1}}

\newcommand*{\affaddr}[1]{#1} 
\newcommand*{\affmark}[1][*]{\textsuperscript{#1}}
\newcommand*\samethanks[1][\value{footnote}]{\footnotemark[#1]}

\usepackage[symbol]{footmisc}
\usepackage[accsupp]{axessibility}

\newcommand{\tikzxmark}{%
\tikz[scale=0.23] {
    \draw[line width=0.7,line cap=round] (0,0) to [bend left=6] (1,1);
    \draw[line width=0.7,line cap=round] (0.2,0.95) to [bend right=3] (0.8,0.05);
}}
\newcommand{\tikzcmark}{%
\tikz[scale=0.23] {
    \draw[line width=0.7,line cap=round] (0.25,0) to [bend left=10] (1,1);
    \draw[line width=0.8,line cap=round] (0,0.35) to [bend right=1] (0.23,0);
}}

\def\paperID{10772} 
\def\confName{CVPR}
\def\confYear{2026}

\title{A Training-Free Style-aligned Image Generation with \\Scale-wise Autoregressive Model}

\author{%
\large
Jihun Park\footnote{}~ \affmark[1], Jongmin Gim\samethanks~ \affmark[1], Kyoungmin Lee\samethanks~ \affmark[1], Minseok Oh \affmark[1], \\
Minwoo Choi \affmark[1], Jaeyeul Kim \affmark[1], Woo Chool Park \affmark[2] and Sunghoon Im\footnote{}~~\affmark[1]\\
\normalsize{\affaddr{\affmark[1]DGIST, Daegu, Republic of Korea}}
\normalsize{\affaddr{\affmark[2] KETI, Republic of Korea}}\\
{\tt\small \{pjh2857, jongmin4422, kyoungmin, harrymark0, subminu, jykim94,  sunghoonim\}@dgist.ac.kr}$^{1}$
 \\
{\tt\small \{wcpark\}@keti.re.kr}$^{2}$
}

\begin{document}

\maketitle

\begin{abstract}
We present a training-free style-aligned image generation method that leverages a scale-wise autoregressive model. While large-scale text-to-image (T2I) models, particularly diffusion-based methods, have demonstrated impressive generation quality, they often suffer from style misalignment across generated image sets and slow inference speeds, limiting their practical usability. To address these issues, we propose three key components: initial feature replacement to ensure consistent RGB statistics, pivotal feature interpolation to align object placement and style, and dynamic style injection, which reinforces style consistency using a schedule function. Unlike previous methods requiring fine-tuning or additional training, our approach maintains fast inference while preserving individual content details. Extensive experiments show that our method achieves generation quality comparable to competing approaches, significantly improves style alignment, and delivers inference speeds over 6$\times$ faster than the fastest style-aligned model.

\end{abstract}

\footnotetext[1]{Equal contribution.}
\footnotetext[2]{Corresponding author.}

\renewcommand*{\thefootnote}{\arabic{footnote}}    
\section{Introduction}
\label{sec:intro}
Large-scale text-to-image (T2I) models \cite{rombach2022highresolution, ramesh2021zeroshot, saharia2022photorealistictexttoimagediffusionmodels, han2024infinity, podell2023sdxlimprovinglatentdiffusion, chang2023musetexttoimagegenerationmasked, tang2024hartefficientvisualgeneration} have become essential tools across various creative fields, including digital content creation, game design, advertising, and artistic visualization.

However, the growing use of these models has revealed a key limitation: style misalignment, where images fail to maintain a consistent visual style across objects, prompting various research efforts \cite{shah2023ziplorasubjectstyleeffectively, frenkel2024implicitstylecontentseparationusing, hertz2024stylealignedimagegeneration, ye2023ipadaptertextcompatibleimage, ruiz2023dreambooth}. One prominent approach leverages parameter-efficient fine-tuning (PEFT), such as Low-Rank Adaptation (LoRA) \cite{hu2021loralowrankadaptationlarge}, which has been applied to specific components of diffusion architectures \cite{shah2023ziplorasubjectstyleeffectively, frenkel2024implicitstylecontentseparationusing}. These methods enable efficient style alignment with minimal computational overhead during training. While effective, these methods still require costly additional fine-tuning. Alternatively, training-free methods have also been proposed \cite{hertz2024stylealignedimagegeneration, zhang2025alignedgen}, such as modifying the self-attention layer with a shared attention layer to enforce style consistency. Despite these efforts, the inherent issue of long inference times in diffusion-based models remains unresolved.

\begin{figure}[t]
    \centering
    \includegraphics[width=.9\columnwidth]{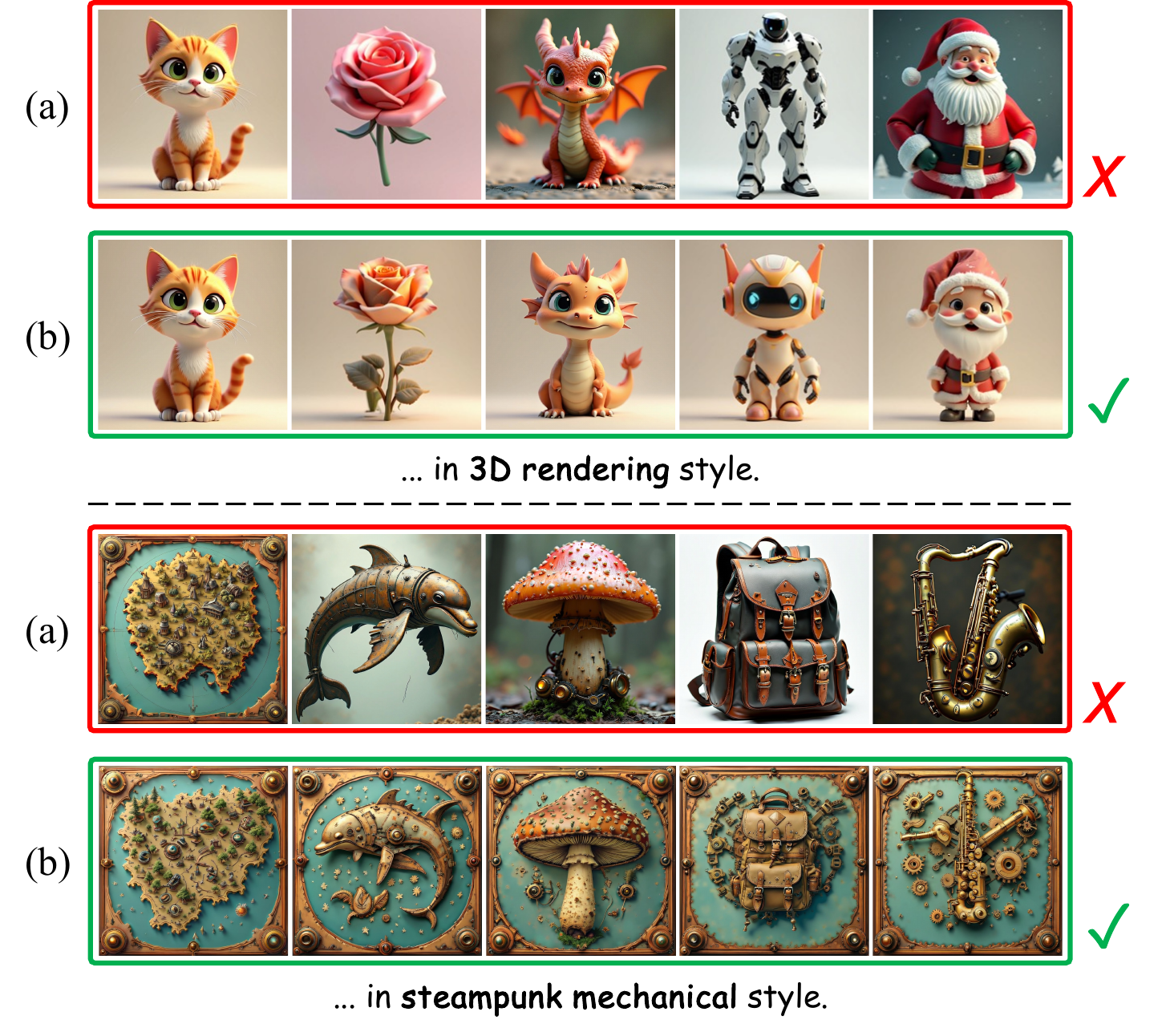}
    \captionof{figure}{Comparison between (a) Standard Text-to-Image model (\textcolor{red}{style misaligned}) and (b) Ours (\textcolor{ForestGreen}{style aligned}). The top rows use the text prompts ``A \{Cat, Rose, Dragon, Robot, Santaclaus\}'' and the bottom rows use ``A \{Map, Dolphin, Mushroom, Backpack, Saxophone\}''.}
    \label{fig:teaser}
\end{figure}

As a faster alternative, vector quantized (VQ)-based autoregressive models \cite{van2017neural, esser2021taming} have been proposed to generate images by predicting discrete tokens. The emergence of mask schedule-based non-autoregressive transformers \cite{chang2022maskgit, kondratyuk2023videopoet, chang2023musetexttoimagegenerationmasked}, inspired by BERT \cite{devlin2019bert}, has further improved both the generation speed and quality of these models. 
Building on this progress, the recently proposed VAR-based next-scale prediction method \cite{tian2024visual} has pushed the performance of autoregressive models even closer to that of diffusion models \cite{tang2024hartefficientvisualgeneration, han2024infinity}, while maintaining significantly faster inference times. 
However, similar to diffusion-based models, autoregressive models also struggle with a style misalignment issue, as illustrated in \figref{fig:teaser}. LoRA-based style fine-tuning has been applied to non-autoregressive transformers \cite{sohn2023styledrop} to address this issue, but it remains computationally expensive and time-consuming due to the additional training required.

To address these challenges, we propose a training-free style-aligned image generation framework that leverages a scale-wise text-to-image (T2I) autoregressive model. 
We begin by conducting a comprehensive analysis of autoregressive model behavior across the generation process as described in \secref{sec:observation}. 
This analysis reveals that the overall RGB statistics and foundational appearance of generated images are predominantly determined during the early-stage, with subsequent scales progressively refining the image using the coarse output from previous steps.
Guided by this observation, we introduce \textit{initial feature replacement} that assigns identical features at the early generation stage to enforce consistent RGB statistics across images, while still allowing each image to maintain its unique content.

\begin{figure}[t]
    \centering
    \includegraphics[width=.85\columnwidth]{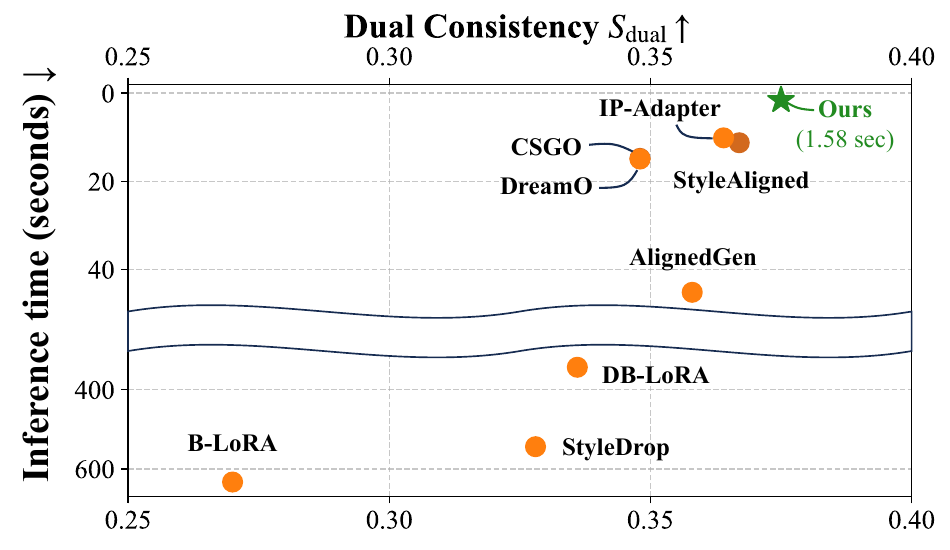}
    \captionof{figure}{Inference time (\(\downarrow\), lower is better) vs. dual consistency (\(\uparrow\), higher is better) curve comparing ours with competitive methods (StyleAligned \cite{hertz2024stylealignedimagegeneration}, B-LoRA \cite{frenkel2024implicitstylecontentseparationusing}, StyleDrop \cite{sohn2023styledrop}, DreamBooth-LoRA (DB-LoRA) \cite{ryu2023low}, IP-Adapter \cite{ye2023ipadaptertextcompatibleimage}, AlignedGen \cite{zhang2025alignedgen}, CSGO \cite{xing2024csgo} and DreamO \cite{mou2025dreamo}).
    }
    \label{fig:analys_infer}
\end{figure}

In addition, we identify that the mid-stage of the coarse-to-fine generation process plays a critical role in determining object placement and the overall visual style of the scene. To guide these aspects, we propose \textit{pivotal feature interpolation}, which smoothly interpolates features during the mid-stage to enforce consistent object positioning and a coherent visual style between images.
Finally, to further enhance style consistency throughout the entire generation process, we introduce \textit{dynamic style injection}. 
This technique employs a schedule function to gradually control the interpolation of self-attention values, striking a balance between maintaining style consistency and preserving each image’s distinct details and content.
By integrating these three components, our approach achieves effective style alignment across images while preserving the high generation quality of the original autoregressive model, all without requiring additional training, as shown in \figref{fig:analys_infer}.

In summary, our primary contributions include:
\begin{itemize}  
    \item We conduct a comprehensive analysis of the scale-wise autoregressive model, revealing the distinct roles of each generation step.
    \item Based on our analysis, we introduce three simple yet effective techniques—\textit{initial feature replacement}, \textit{pivotal feature interpolation}, and \textit{dynamic style injection}—to enhance style consistency while preserving content fidelity.
    \item To the best of our knowledge, this is the first work to propose \textit{a training-free style-aligned image generation} based on an autoregressive model, achieving at least 6$\times$ faster inference along with the best style alignment performance.
\end{itemize}

\section{Related works}
\label{sec:related}

\begin{figure*}[t]
    \centering
    \includegraphics[width=.80\textwidth]{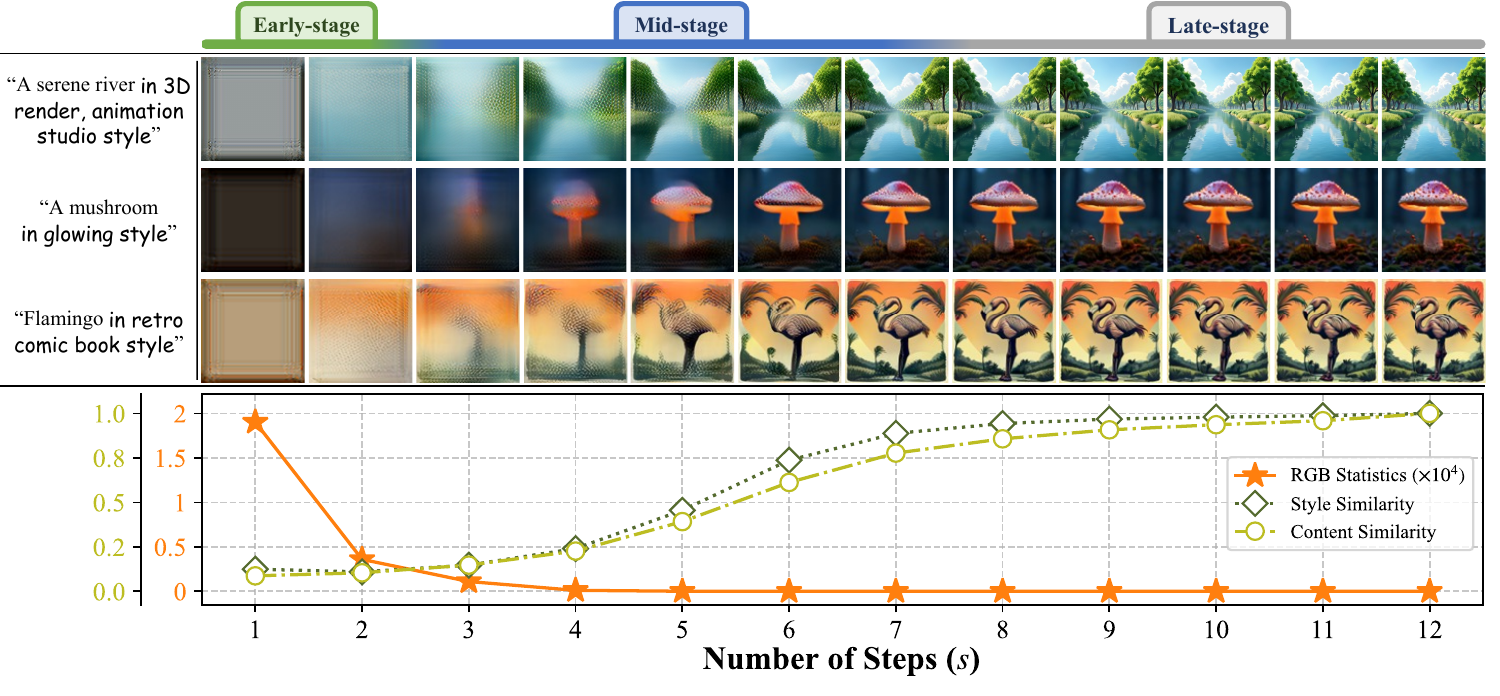}
    \captionof{figure}{\textbf{Visualization of images generated at different steps of the next-scale prediction process.} In the early- and mid-stages, global composition and overall style are established, while later steps focus on refining details and textures. We also track RGB statistics, content similarity, and style similarity across 400 generated images, comparing each step to the final output (12th step) to evaluate the progression of content preservation, style consistency, and RGB statistics.}
    \label{fig:analys}
\end{figure*}

\subsection{Text-to-image generation}
The development of large-scale text-image pair datasets \cite{schuhmann2022laion5bopenlargescaledataset, kakaobrain2022coyo-700m, changpinyo2021conceptual12mpushingwebscale, lin2015microsoft}, combined with advances in generative models such as diffusion models \cite{ho2020denoising, song2022denoising}, GANs \cite{goodfellow2014generative}, and autoregressive models \cite{van2017neural, esser2021taming}, has significantly accelerated the progress of large-scale text-to-image (T2I) generation models \cite{rombach2022highresolution, ramesh2021zeroshot, saharia2022photorealistictexttoimagediffusionmodels, han2024infinity, podell2023sdxlimprovinglatentdiffusion, chang2023musetexttoimagegenerationmasked, tang2024hartefficientvisualgeneration, kang2023gigagan}. 
In particular, diffusion-based text-to-image models have demonstrated outstanding performance, making them widely adopted for various downstream applications such as style transfer and image editing \cite{chung2024style, tumanyan2022plugandplay, brooks2023instructpix2pix, jeong2024trainingfree}. More recently, autoregressive models employing next-scale prediction \cite{tian2024visual} have emerged as promising alternatives, offering significantly faster inference speeds while maintaining competitive generation quality compared to diffusion-based approaches. 
This advancement has opened new avenues for efficient T2I generation \cite{han2024infinity, tang2024hartefficientvisualgeneration}. 
Despite these advancements, both diffusion and autoregressive models still suffer from persistent challenges in style alignment, which limits their practical usability and degrades user experience.

\subsection{Style Transfer}
Image style transfer, pioneered by \cite{Gatys_2016_CVPR}, leverages pre-trained CNNs like VGGNet to extract content and style features, laying the groundwork for this research area. 
However, these early optimization-based methods suffer from high computational costs, as they require per-image optimization. 
To mitigate this, Huang \textit{et al.} \cite{Huang_2017_ICCV} introduces Adaptive Instance Normalization (AdaIN) for real-time style transfer, followed by Whitening and Coloring Transform (WCT) \cite{li2017universal, li2018closed}, which improves feature alignment. 
With the rise of attention mechanisms \cite{vaswani2023attention, dosovitskiy2021image}, newer models \cite{liu2021adaattn, hong2023aespanet, yao2019attentionaware} have further enhanced stylization quality. 
Generative models such as GANs \cite{goodfellow2014generative}, diffusion models \cite{ho2020denoising}, and vector quantization-based autoregressive models \cite{esser2021taming} have opened new possibilities in style transfer. 
Diffusion-based methods \cite{chung2024style, yang2023zero, kwon2023diffusionbased} achieve high visual fidelity by integrating style features through cross-attention during the denoising process, while vector quantization approaches \cite{chen2022vector, xu2023stylerdalle, huang2023quantart, Gim_2024_ACCV} enhance stylization by combining content tokenization with learned style representations. 
Despite their advantages, early vector quantization models often require separate style codebooks or extensive fine-tuning, while diffusion-based methods face challenges due to their long inference times.

\subsection{Personalized image generation}
Various personalized image generation methods have been proposed to adapt new visual features to user intent using pre-trained text-to-image models. These methods can be broadly categorized into content-oriented and style-oriented methods.
Content-oriented methods \cite{ruiz2023dreambooth, li2023blip, wei2023elite} primarily focus on capturing object-specific attributes or synthetic features to generate images that explicitly reflect a target subject. They leverage pre-trained models or fine-tuning methods to embed subject-specific characteristics into the generated images.
In contrast, style-oriented methods~\cite{sohn2023styledrop, frenkel2024implicitstylecontentseparationusing, ryu2023low, zhang2025alignedgen, xing2024csgo, mou2025dreamo} focus on controlling the style of the generated images. In particular, Style-Aligned Generation \cite{hertz2024stylealignedimagegeneration}, which closely relates to our approach, enhances style-consistency across batches by sharing attention during the denoising diffusion process and reducing differences among generated images through AdaIN.

However, these methods generally rely on diffusion-based generation or require fine-tuning, resulting in high computational costs and longer processing times. In contrast, we propose a training-free, scale-wise autoregressive model that achieves style-consistent image generation with significantly reduced inference time.
\section{Method}
\label{sec:method}

\begin{figure*}[t!]
    \centering    \includegraphics[width=.8\textwidth]{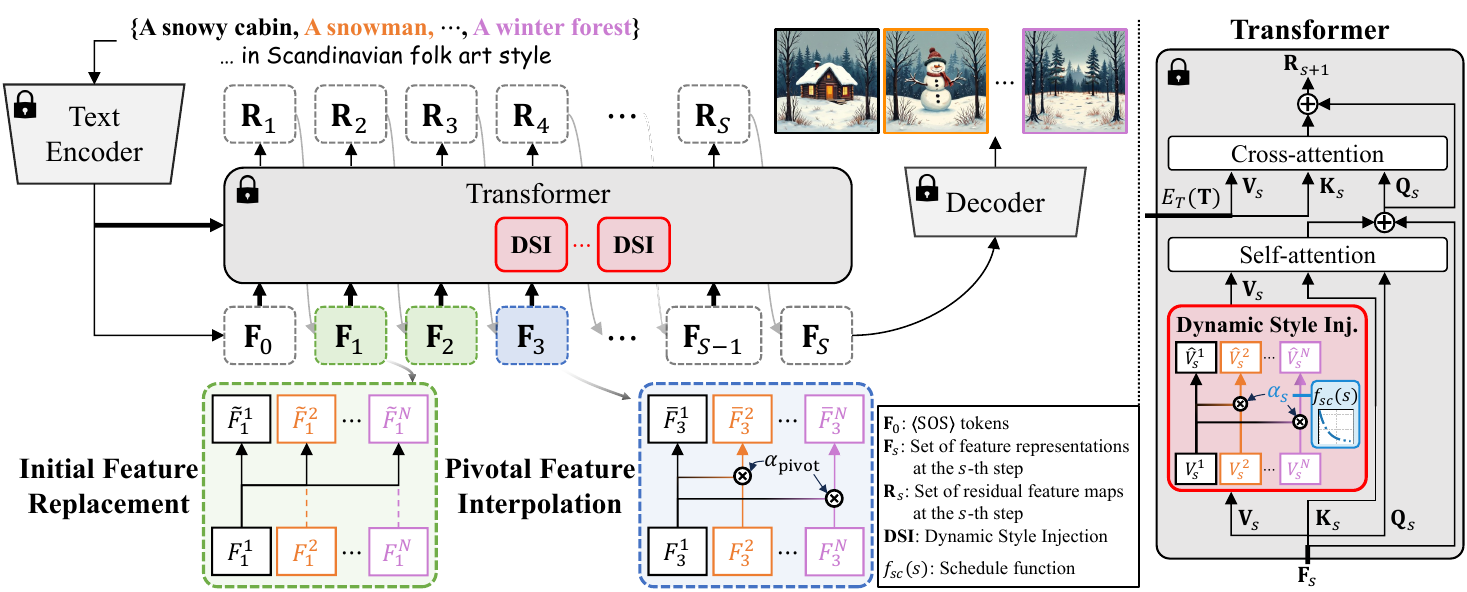}
    \captionof{figure}{\textbf{Overall pipeline of our model.} The T5 text encoder processes text prompts $\mathbf{T}$, providing conditions and $\langle \text{SOS} \rangle$ tokens to the transformer. \textit{Initial Feature Replacement} aligns RGB statistics at the 1st and 2nd generation steps. \textit{Pivotal Feature Interpolation} adjusts object positions and styles at the \(\bar{s}\)-th step (\(\bar{s}\)=3), while \textit{Dynamic Style Injection} gradually reduces style influence from the 3rd to 7th steps. The decoder converts the transformer’s final step output $\mathbf{F}_{S}$ into the style-aligned images $\mathbf{I}$.}
    \label{fig:overall}
\end{figure*}

\subsection{Observation of next-scale prediction}
\label{sec:observation}
To gain deeper insight into the internal mechanisms underlying the next-scale prediction scheme, we analyze the evolution of RGB statistics, style, and content similarity throughout the autoregressive generation process, as illustrated in \figref{fig:analys}. 
We evaluate the progression of content preservation, style consistency, and RGB statistics by comparing each generation step to the final output (the 12th step) using 400 images generated from randomly sampled text prompts created with ChatGPT. 
RGB statistics are quantified using the chi-square distance between histograms, content similarity with the cosine similarity of VGG19 features \cite{simonyan2014very}, and style similarity with the average pairwise cosine similarity of DINO ViT-B/8 \cite{caron2021emerging} embeddings within each set.
Based on this analysis, we identify the following key observations, which form the foundation of our method.

\noindent \textbf{(1) Dominance of RGB statistics in early-stage:} As shown in \figref{fig:analys}, the early-stage (\textit{e.g.}, the 1st-2nd steps) primarily establishes the overall RGB statistics of the generated images.
Once the RGB statistics are set, the autoregressive model, heavily conditioned on the outputs from previous steps, closely follows these statistics, strongly influencing all subsequent stages.
This means that any modifications made during the early-stage have a significant impact on the final RGB statistics.
This observation highlights the importance of controlling early-stage representations to ensure consistent style across a set of generated images.
Based on this observation, we propose the initial feature replacement method described in \secref{sec:initial}, enforcing consistent RGB statistics during the early-stage to promote style alignment across generated images.

\noindent \textbf{(2) Dynamic transition of style and content in mid-stage:} \figref{fig:analys} further reveals that both style and content gradually evolve with increasing detail during the mid-stage, eventually stabilizing as the generation process approaches the later steps (\textit{e.g.}, the 8th step). Due to the autoregressive nature of the model, early-stage decisions strongly influence the overall generation flow, shaping both style and content composition throughout the process. This observation aligns with findings from \cite{voronov2024swittidesigningscalewisetransformers}, demonstrating that modifications in the late-stage have minimal impact on the overall semantic structure, while adjustments made during the mid-stage effectively blend both style and content attributes. Based on this, we identify the mid-stage as an optimal window for style refinement, enabling us to enhance style consistency without significantly altering the underlying content structure. This insight serves as the foundation of our strategy to apply targeted interventions during these intermediate steps, harmonizing style and content before they fully stabilize in the late-stage.

\noindent \textbf{(3) Quantitative analysis of feature dynamics:}  
To further substantiate these observations, we conduct a quantitative analysis of the scale-wise generation process, focusing on feature variations across early, mid, and late-stages. Specifically, we measure the average differences between consecutive generation steps in both content and style features. For each region of the generation process—early (steps 1–2), mid (steps 3–7), and late (steps 8–12)—we compute the mean feature changes as follows:
\begin{itemize}
    \item \textbf{Content feature differences:} Early: 0.031, Mid: 0.137, Late: 0.042
    \item \textbf{Style feature differences:} Early: 0.002, Mid: 0.158, Late: 0.018
\end{itemize}
These statistics quantitatively confirm that both content and style features vary most significantly during the mid-stage, validating our stage-wise decomposition of the generation process into early, mid, and late phases.

To leverage this observation, we propose the pivotal feature interpolation described in \secref{sec:attention}, which aligns the object’s silhouette and ensures uniform style by adjusting features at one of the steps in mid-stage.
In addition, we introduce the dynamic style injection in \secref{sec:style_injection}, which dynamically injects style using a designed scheduling function. This ensures that style consistency is continuously reinforced throughout the generation process, while allowing content details to be progressively refined without distortion.

\subsection{Overall pipeline} 
\label{sec:overall}
In this paper, we aim to generate a set of images \(\mathbf{I}=\{I^n\}_{n=1}^N\) corresponding to an input set of text prompts \(\mathbf{T}=\{T^n\}_{n=1}^N\), while ensuring that all generated images share a consistent visual style. To achieve this, we concatenate the text prompts and process them in parallel as a batch, enabling the model to generate multiple images simultaneously with shared stylistic coherence.
The overall pipeline of our model is illustrated in \figref{fig:overall}. Our approach builds upon the architecture of Infinity \cite{han2024infinity}, a state-of-the-art T2I model based on the next-scale prediction paradigm \cite{tian2024visual}. Specifically, our model consists of a pre-trained text encoder \(E_T\) derived from Flan-T5 \cite{chung2022scalinginstructionfinetunedlanguagemodels}, an autoregressive transformer model \(M\) responsible for next-scale prediction, and a decoder \(D\) that reconstructs the final image from the predicted residual feature maps. 
Based on our observations, we define the set of all generation steps as $\mathbf{S}=\{1,2,\dots,S\}$, the early steps as $\mathbf{S}_\text{e}=\{1,2\}$, and middle steps as $\mathbf{S}_\text{m}=\{3,\dots,7\}$. 

The autoregressive model $M$ predicts a sequence of quantized residual feature maps \( \mathbf{R}_s = \{R^n_s\}_{n=1}^{N}\), conditioned on the input text prompts \( \mathbf{T}\) and previously generated features $\mathbf{F}_{s-1}$. The initial features $\mathbf{F}_0$ correspond to the start-of-sequence \( \langle \text{SOS} \rangle \) tokens. The prediction process at all generation steps $s\in \mathbf{S}$ can be described as follows:
\begin{gather}
\begin{split}
    \mathbf{R}_s &= M(\mathbf{F}_{s-1}, E_T(\mathbf{T})) \\
   & = M_{CA}(M_{SA}(\mathbf{Q}_{s-1}, \mathbf{K}_{s-1}, \mathbf{V}_{s-1}), E_T(\mathbf{T})),
\end{split}
\end{gather}
where $\mathbf{Q}$, $\mathbf{K}$ and $\mathbf{V}$ denote the query, key and value obtained by multiplying $\mathbf{F}$ with
$\mathbf{W}_{Q}$, $\mathbf{W}_{K}$ and $\mathbf{W}_{V}$, respectively.
\(M_{SA}(\cdot)\) and \(M_{CA}(\cdot)\) denote self-attention and cross-attention within transformer model \(M\), respectively.

At each step, the predicted residual feature maps $\mathbf{R}_s$ are upsampled to $H \times W$ using a bilinear upsampling function \(\text{up}_{H\times W}(\cdot)\) and summed to serve as inputs to the autoregressive model for the next scale as follows:
\begin{equation}
    \mathbf{F}_s = \sum_{i=1}^{s} \text{up}_{H\times W}(\mathbf{R}_i),~~\mathbf{R}_s \in \mathbb{R}^{N \times h_s \times w_s} \\
\end{equation}
where $h_s$ and $w_s$ are the size of the residual feature at generation step $s$. 
Finally, the output image set $\mathbf{I}$ is generated passing the complete set of feature representations $\mathbf{F}_S$ at the last step $S$ through the decoder $D$ as follows:
\begin{equation}
\begin{gathered}
    \mathbf{I} = D(\mathbf{F}_S).
\end{gathered}
\end{equation}

\subsection{Initial feature replacement}
\label{sec:initial}
As described in \secref{sec:observation}, the initial steps in next-scale prediction play a dominant role in establishing the overall RGB statistics of the generated images, while contributing minimally to fine-grained content. This observation aligns closely with the core design philosophy of next-scale prediction in VAR \cite{tian2024visual}, where the multi-scale, coarse-to-fine process naturally enforces a structured ``order'' on image generation.  

Motivated by this strong correlation, we adopt a simple yet effective initialization strategy. All features from $N$ images at the early generation steps in $\mathbf{S}_{\text{e}}$, denoted as $\mathbf{F}_{s}=\{F^n_{s}\}^N_{n=1}$, are replaced with the first feature of the batch $F_{s}^1$ as follows:

\begin{equation}
\begin{gathered}
        \mathbf{F}_{s}\leftarrow\{\tilde{F}_{s}^{n}\}_{n=1}^N,~\forall{s}\in\mathbf{S}_{\text{e}},\\
        \tilde{F}_{s}^{n} = F_{s}^{1},~\forall n \in \mathbf{N},\\ 
\end{gathered}
\end{equation}
where $\mathbf{N} = \{1,2, \dots, N\}$ denotes the set of image indices.
This replacement helps ensure that the images generated in later steps within a batch share consistent and harmonious RGB statistics, while still preserving the unique content and identity of each individual image.

\subsection{Pivotal feature interpolation}
\label{sec:attention}
The style-aligned text-to-image generation task \cite{hertz2024stylealignedimagegeneration}  aims to produce a set of images that maintain consistent object placement and a uniform style across all generated images. Based on our observations in \secref{sec:observation}, we identify that one of the mid-stage steps in the coarse-to-fine next-scale prediction process plays a key role in determining both object placement and overall style. To leverage this property, we apply feature interpolation at $\bar{s}$-th step within the mid-stage $\mathbf{S}_{\text{m}}$, where each feature is interpolated with the first feature in the batch. The feature interpolation is defined as:
\begin{equation}
\begin{gathered}
        \mathbf{F}_{\bar{s}}\leftarrow\{\bar{F}_{\bar{s}}^{n}\}_{n=1}^N,~\bar{s} \in \mathbf{S}_{\text{m}},\\
        ~\bar{F}_{\bar{s}}^{n} = \alpha_{\text{pivot}}{F}_{\bar{s}}^{1} + (1-\alpha_{\text{pivot}}){F}_{\bar{s}}^{n},~\forall n \in \mathbf{N},
\end{gathered}
\end{equation}
where \(\alpha_{\text{pivot}}\) denotes interpolation weight. This guides the generation process, ensuring consistent object placement and visual style across the generated images.

\subsection{Dynamic style injection}
\label{sec:style_injection}
Although pivotal feature interpolation helps guide the generated images toward aligned styles and consistent object placement, it alone is insufficient to fully enforce a uniform style across the entire image set. To address this limitation, and based on our analysis in \secref{sec:observation}, we introduce an enhanced style injection method that updates the value representations \(\mathbf{V}_s=\{V_{s}^{n}\}_{n=1}^N\) within the self-attention module before content and style are fully established.
Our dynamic style injection interpolates the value features of the first image in the batch into the rest, updating the value $\mathbf{V}_s$ at generation step $s$ as follows:

\begin{equation}
\label{eq:style_feature}
    \begin{gathered}
     \mathbf{V}_s\leftarrow\{\hat{V}_s^n\}_{n=1}^N,~\forall s \in \mathbf{S}_{\text{m}},\\
   \hat{V}^n_s = \alpha_s {V}^1_s + (1-\alpha_s){V}^n_s, ~\forall n \in \mathbf{N},
    \end{gathered}
\end{equation}
where the updated $\mathbf{V}_s$ serves as the input to the self-attention module $M_{SA}(\cdot)$. 

The interpolation weight \( \alpha_s \) is dynamically adjusted using a non-linear decreasing schedule function \( f_{sc} \), inspired by the concave increasing trend observed during the mid-stage (steps 3–7) of the style-content similarity graph in \figref{fig:analys}, which follows an overall S-curve pattern. 
This design encourages stronger style injection at the beginning of the mid-stage—when global style attributes are still forming—and gradually weakens it toward the later steps, allowing content features to stabilize and refine naturally. 
The schedule function is defined as:

\begin{equation}
\label{eq:style_feature1}
     \alpha_s=f_{sc} (s) = \frac{e^{-r \cdot \frac{s}{12}} - e^{-r}}{1 - e^{-r}}
\end{equation}
where $r$ denotes the decay rate of the schedule function. 
This style injection process promotes style consistency across the image set while preserving the unique content of each individual image.

\begin{table*}[h]
\caption{Quantitative comparison with state-of-the-art style-aligned image generation models. We evaluate the generated image sets in terms of object relevancy (CLIP text score), style consistency (DINO embedding similarity), and dual consistency (harmonic mean of the CLIP score and DINO embedding similarity). The inference time is measured per image, where for DB-LoRA, B-LoRA, and StyleDrop, the reported time includes both inference and per-style optimization. The symbols $\uparrow$ and $\downarrow$ indicate that higher values are better and lower values are better, respectively.}
\centering
\footnotesize
\begin{tabular}
{c|c|cccccccc}
\toprule
Metric  & \multicolumn{1}{c}{\textbf{Ours}} & \multicolumn{1}{|c}{StyleAligned} & \multicolumn{1}{c}{B-LoRA } & \multicolumn{1}{c}{StyleDrop}  & \multicolumn{1}{c}{DB-LoRA } & \multicolumn{1}{c}{IP-Adapter} & \multicolumn{1}{c}{CSGO}& \multicolumn{1}{c}{AlignedGen}&\multicolumn{1}{c}{DreamO} \\
\midrule
Training-free &\tikzcmark&\tikzcmark&\tikzxmark&\tikzxmark&\tikzxmark&\tikzxmark&\tikzxmark&\tikzcmark&\tikzxmark\\
\midrule
   Object relevancy (\(S_{\text{obj}}\)) $\uparrow$ & {0.282} & 0.281 & \underline{0.302} & {0.267} & \textbf{0.309}  & 0.277 & 0.292 & 0.283& 0.292\\
  Style consistency (\(S_{\text{sty}}\)) $\uparrow$ &  \textbf{0.556} & \underline{0.530} & 0.292 & {0.425} &  0.369  & 0.529 & 0.432 & 0.486& 0.432\\
  Dual Consistency (\( S_{\text{dual}}\)) $\uparrow$ &  \textbf{0.375} & \underline{0.367} & 0.270 & {0.328} & 0.336  & 0.364 & 0.348 & 0.358& 0.348\\
\midrule
   Inference time (seconds) $\downarrow$ &  \textbf{1.58} & 11.25 & 633.20 & 544.06 & 343.68 & \underline{10.14} & 14.80 & 45.21 &14.95\\ 
\bottomrule
\end{tabular}
\label{tab:comparison}
\end{table*}

\begin{figure*}[h]
    \centering
    \includegraphics[width=.9 \linewidth]{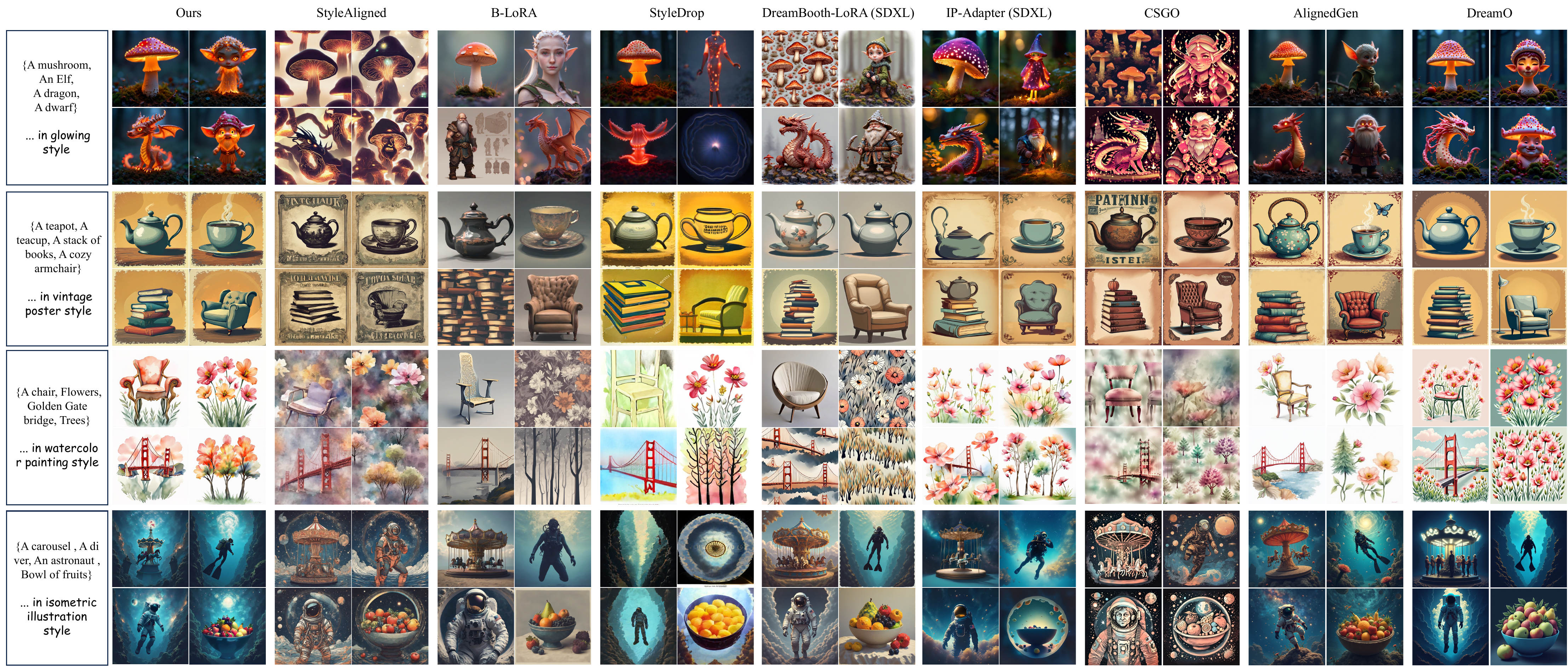}
    \captionof{figure}{Qualitative comparison with state-of-the-art style-aligned image generation models. 
    }
    \label{fig:comparison}
\end{figure*}

\section{Experiments}
\label{sec:experiment}
\subsection{Implementation Details}
We implement our method using pre-trained Infinity~\cite{han2024infinity} 2B model with all parameters frozen, which performs scale-wise prediction across 12 steps. 
Initial feature replacement is applied at all early steps $\mathbf{S}_{\text{e}}$ (see \secref{sec:overall}), followed by pivotal feature interpolation at the 3rd scale ($\bar{s}=3$) with a fixed interpolation weight of $\alpha_{\text{pivot}} = 0.4$. 
Alternatively, $\bar{s}$ can also be applied at any mid-stage scale in $\mathbf{S}_{\text{m}}$ as a hyperparameter, with minimal performance variation (see supplementary material), demonstrating the robustness of our model.
Additionally, dynamic style injection is applied at all mid-stage $\mathbf{S}_{\text{m}}$, where the interpolation weight $\alpha_{s}$ is determined by a schedule function with decay rate $r$ set to 3.4.
Our baseline uses a codebook with a dimension of $2^{32}$, and the quantized feature map has a resolution of 64$\times$64$\times$32. The number of generated style-aligned images can be controlled by specifying the number of input text prompts. Generating a set of four 1024$\times$1024 images simultaneously takes approximately 6.3 seconds (1.58 seconds per image) on a single A6000 GPU.

\subsection{Evaluation metrics}
Following \cite{hertz2024stylealignedimagegeneration}, we adopt the same evaluation methodology to assess the generated image sets based on two key aspects: \textbf{object relevancy} (\(S_\text{obj}\)) and \textbf{style consistency} (\(S_\text{sty}\)).  
To evaluate object relevancy, we compute the CLIP cosine similarity \cite{radford2021learning} between each generated image and its corresponding text description, ensuring that the generated content aligns with the given prompt.
To evaluate style consistency, we utilize DINO ViT-B/8 \cite{caron2021emerging}, which is sensitive to style variations, following the approach of \cite{hertz2024stylealignedimagegeneration}. We then measure the pairwise average cosine similarity of the DINO embeddings within each generated set.
Additionally, we propose a combined metric, \textbf{dual consistency} (\(S_\text{dual}\)), to provide a balanced assessment of both object relevancy and style consistency.
We compute it as the harmonic mean of the two metrics $S_{\text{dual}} = 2 S_{\text{obj}} S_{\text{sty}}/\big({S_{\text{obj}} + S_{\text{sty}}}\big)$.
For a fair comparison, we adopt the same 100 text prompts generated by ChatGPT as used in \cite{hertz2024stylealignedimagegeneration}, generating 4 image sets per prompt and totaling 400 images for evaluation.

\subsection{Comparison with state-of-the-art style-aligned image generation models}
\noindent\textbf{Quantitative comparison.} 
\tabref{tab:comparison} presents a quantitative comparison between our method and state-of-the-art style-aligned image generation models: StyleAligned \cite{hertz2024stylealignedimagegeneration}, B-LoRA \cite{frenkel2024implicitstylecontentseparationusing}, StyleDrop \cite{sohn2023styledrop}, DreamBooth-LoRA (DB-LoRA) \cite{ryu2023low}, IP-Adapter \cite{ye2023ipadaptertextcompatibleimage}, CSGO \cite{xing2024csgo}, AlignedGen \cite{zhang2025alignedgen}, and DreamO \cite{mou2025dreamo}. The results demonstrate that our method achieves the best balance between object relevancy and style consistency, while also significantly outperforming all compared models in inference speed.

Our method achieves comparable object relevancy to StyleAligned~\cite{hertz2024stylealignedimagegeneration} and AlignedGen~\cite{zhang2025alignedgen}, while significantly outperforming both in style and dual consistency. Moreover, it is approximately $6$--$28\times$ faster ($1.58$\,s vs.\ $11.25$\,s and $45.21$\,s per image), demonstrating the superior efficiency of our framework.
Although our object relevancy score is slightly lower than B-LoRA \cite{frenkel2024implicitstylecontentseparationusing} and DB-LoRA \cite{ryu2023low}, our method achieves substantially higher style consistency, leading to the highest score in the dual consistency metric.
Furthermore, our method is approximately $400\times$ faster than B-LoRA \cite{frenkel2024implicitstylecontentseparationusing} and $200\times$ faster than DB-LoRA \cite{ryu2023low}, highlighting its exceptional efficiency.

\noindent\textbf{Qualitative comparison.}  
To further demonstrate the effectiveness of our proposed method, we present a qualitative comparison with existing models in \figref{fig:comparison}. 
B-LoRA~\cite{frenkel2024implicitstylecontentseparationusing} and DB-LoRA~\cite{ryu2023low} exhibit strong object fidelity owing to their high object relevancy, yet their results reveal a clear lack of style consistency compared to our model and other training-free approaches. 
StyleAligned~\cite{hertz2024stylealignedimagegeneration} and AlignedGen \cite{zhang2025alignedgen} produce results visually comparable to ours, but subtle differences remain: in StyleAligned, the generated objects are occasionally incomplete or distorted (e.g., in the first and second rows), while AlignedGen shows relatively inconsistent style coherence across images (e.g., in the second and third rows). 
In contrast, the results show that our method consistently preserves both object structure and stylistic uniformity. These qualitative results demonstrate that our approach achieves a more stable balance between object fidelity and style consistency, leading to high-quality, style-aligned generations.

\begin{table}[t]
    \caption{Ablation study on the initial feature replacement (Re), Pivotal Feature Interpolation (PFI), Dynamic style injection (DSI). The symbol $\uparrow$ indicates that higher values are better.}
    \centering
    \footnotesize
    \resizebox{0.8\linewidth}{!}{
    \begin{tabular}{c|c c c|c|c c}
    \toprule
      \multicolumn{1}{c|}{} & \multicolumn{3}{c|}{Component} & \multicolumn{3}{c}{Quantitative Metrics} \\
      \# &Re & PFI & DSI & { $S_{\text{dual}}$ $\uparrow$} & {$S_{\text{obj}}$ $\uparrow$} & {$S_{\text{sty}}$ $\uparrow$} \\
    \midrule
     (a)&  &  &  &   0.296 & \textbf{0.298} & 0.295 \\
     (b)&\checkmark &  &  & 0.327 & \underline{0.294} & 0.368 \\
     (c)&\checkmark & \checkmark &  & \underline{0.337} & 0.292 & \underline{0.397} \\
     (d)&\checkmark & \checkmark & \checkmark & \textbf{0.375} & 0.282 & \textbf{0.556} \\
    \bottomrule
    \end{tabular}    
    }
    \label{tab:ablation}
\end{table}

\begin{figure}[t]
    \centering
    \includegraphics[width=.85 \linewidth]{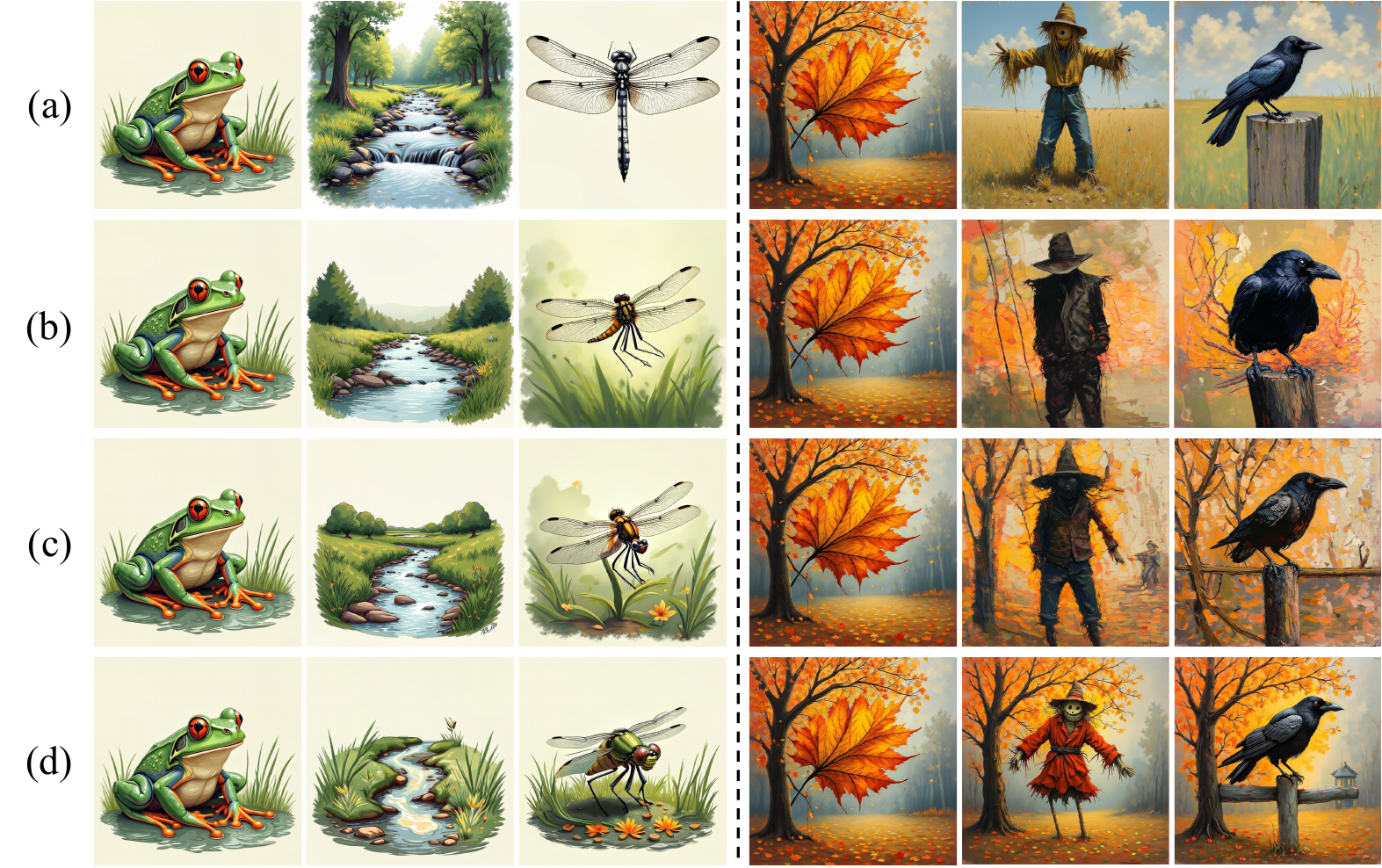}
    \captionof{figure}{Qualitative analysis of ablation study. The results from (a)-(d) correspond to the configurations in \tabref{tab:ablation}.}
    \label{fig:ablation}
\end{figure}

\subsection{Ablation study}
\noindent\textbf{Quantitative analysis.}  
The quantitative results in \tabref{tab:ablation} validate the contributions of each proposed technique to improving both style consistency and overall image quality. Each component offers distinct benefits. As shown in \tabref{tab:ablation}-(b), \textit{Initial feature replacement} enhances style consistency by ensuring uniform RGB statistics, leading to a clear increase in \(S_{\text{sty}}\). In \tabref{tab:ablation}-(c), \textit{Pivotal feature interpolation} further improves style alignment while preserving object relevancy. Integrating all components including \textit{dynamic style injection}, as shown in \tabref{tab:ablation}-(d), significantly boosts style consistency results while resulting in a slight decrease in object relevancy. However, this minor trade-off is outweighed by substantial gains in style consistency and dual consistency. These results confirm that each module effectively contributes to improving style alignment while preserving content fidelity across the generated image sets.

\noindent\textbf{Qualitative analysis.} The qualitative results also reinforce these findings. As seen in \figref{fig:ablation}-(a), the absence of our proposed techniques results in misaligned styles across images, with noticeable variations in color tones and lighting conditions. \figref{fig:ablation}-(b), incorporating \textit{Initial feature replacement}, produces more uniform background colors, improving overall style consistency. In \figref{fig:ablation}-(c), where \textit{Pivotal feature interpolation} is additionally applied, the generated images exhibit enhanced coherence in object placement and style alignment. Finally, \figref{fig:ablation}-(d) integrates all three proposed techniques, including \textit{Dynamic style injection}, which further refines style consistency while preserving content structure. 
These qualitative ablation results demonstrate that each module is crucial in improving style alignment while maintaining object fidelity across the generated image sets.

\subsection{User study}
We conduct a user study to further enhance our evaluation, with the results presented in \tabref{tab:user_study}. The study involved 50 participants, ranging in age from their 20s to 50s. Participants were asked to evaluate two key aspects: \textit{object relevancy} and \textit{style consistency}.
We compare images generated by our model with those produced by StyleAligned \cite{hertz2024stylealignedimagegeneration}, AlignedGen \cite{zhang2025alignedgen}, and IP-Adapter \cite{ye2023ipadaptertextcompatibleimage}, the top three performers in quantitative evaluations for style-aligned image generation.
The user study results demonstrate that our model effectively preserves object relevancy while achieving strong style consistency across the generated image sets.

\begin{table}[t]
\caption{User study detailing preference percentages.}
\small
\centering
\footnotesize
\begin{tabular}{ccc}
\toprule
  Method&  Object  $\uparrow$ & Style  $\uparrow$ \\ \midrule
StyleAligned \cite{hertz2024stylealignedimagegeneration} & 16.9 \% & 10.9 \%  \\
AlignedGen \cite{zhang2025alignedgen}   & 29.7 \% & 32.8 \%\\
IP-Adapter \cite{ye2023ipadaptertextcompatibleimage}  & 8.0\% & 17.8 \%\\
\midrule
\textbf{Ours}      & \textbf{45.4} \% & \textbf{38.5} \%\\ \bottomrule
\end{tabular}
\label{tab:user_study}
\end{table}

\begin{figure}[t]
    \centering
    \includegraphics[width=0.8 \linewidth]{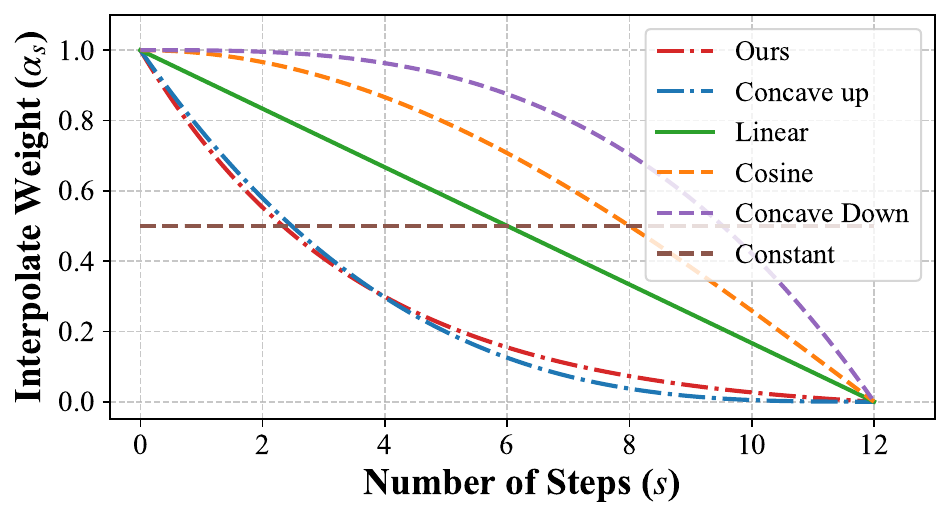}
    \captionof{figure}{Visualization of various schedule functions.}
    \label{fig:schedule}
\end{figure}

\begin{table}[t]
    \caption{Additional ablation study on various schedule functions.}
    \centering
    \footnotesize
    \resizebox{0.7\linewidth}{!}{
    \begin{tabular}{c|c|cc}
    \toprule
       \multicolumn{1}{c|}{Schedule} & \multicolumn{3}{c}{Quantitative Metrics} \\ {function} &
       {$S_{\text{dual}}$ $\uparrow$} &
      {$S_{\text{obj}}$ $\uparrow$} & {$S_{\text{sty}}$ $\uparrow$}  
      \\
    \midrule
     Constant & 0.359 & 0.223 & 0.929 \\
     Linear & 0.360 & 0.223 & 0.936 \\
     Concave Up & \underline{0.369} & \textbf{0.283} & 0.531 \\
     Concave Down & 0.360 & 0.222 & \textbf{0.954} \\
     Cosine & 0.361 & 0.223 & \underline{0.942} \\
    \midrule
    \textbf{Ours} \(f_{sc}(s)\) & \textbf{0.375} & \underline{0.282} & 0.556 \\
    \bottomrule
    \end{tabular}    }
    \label{tab:add_ablation}
\end{table}

\subsection{Schedule function design}
Based on our observations in \figref{fig:analys}, we hypothesize that gradually incorporating style information during the stages where style and content are progressively defined would enhance style alignment while preserving the intended content structure. To validate this hypothesis, we conduct an additional ablation study by designing five distinct schedule functions(illustrated in \figref{fig:schedule}) that progressively decay over the generation steps and evaluate their impact.

As shown in \tabref{tab:add_ablation}, our proposed schedule function, derived directly from our observations, achieves the highest \( S_{\text{dual}} \) score, demonstrating its effectiveness in balancing both object relevancy and style consistency. Notably, the concave up schedule, which follows a decay pattern similar to our design, also performs relatively well, further supporting our hypothesis. In contrast, the linear, constant, concave-down, and cosine schedules, which deviate from our observed trends, exhibit lower performance. These results confirm that our observation-driven schedule function optimally integrates style while preserving content.

\subsection{Incorporating ours to other models} 
To demonstrate the generalization ability of our method, we applied it to alternative weights of the baseline models, Infinity 8B, and to another T2I model using scale-wise autoregressive generation, Switti \cite{voronov2024swittidesigningscalewisetransformers}. As shown in \tabref{tab:ablation2}, our method significantly improves $S_\text{dual}$, indicating effective style alignment while maintaining object fidelity. Notably, the autoregressive model Switti demonstrates consistent style alignment despite having a different architecture from our baseline. These results support the generality of our approach across varying model configurations.

\begin{table}[t]
    \caption{Incorporating ours to other scale-wise autoregressive-based Text-to-Image models.}
    \centering
    \footnotesize
    \resizebox{0.7\linewidth}{!}{
    \begin{tabular}{lccc}
    \toprule
    Method  & \( S_{\text{dual}}\) $\uparrow$ & \(S_{\text{obj}}\) $\uparrow$ & \(S_{\text{sty}}\) $\uparrow$ \\
    \midrule
    Switti & 0.308 & \textbf{0.294} & 0.323  \\
    Switti + Ours & \textbf{0.365} & 0.281 & \textbf{0.521}  \\
    \midrule
    Infinity 8B & 0.316 & \textbf{0.295} & 0.340  \\
    Infinity 8B + Ours & \textbf{0.373} & 0.274 & \textbf{0.585}  \\
    \bottomrule
    \end{tabular}    }

    \label{tab:ablation2}
\end{table}




\section{Conclusion}
\label{sec:conclusion}
In this paper, we introduce a practical and efficient training-free style-aligned image generation method built on a scale-wise autoregressive model. To this end, we analyze the next-scale prediction scheme in terms of color statistics, content evolution, and style variation across generation steps. We observe that the early-stage primarily determines the overall RGB statistics, forming the foundation for the final appearance, while the mid-stage progressively refines both content and style to achieve visual coherence.

Building on these insights, we propose three components to enhance style alignment and content preservation. \textit{Initial feature replacement} ensures consistent RGB statistics by stabilizing early-stage features. \textit{Pivotal feature interpolation} improves object placement and spatial coherence at critical mid-stage steps. \textit{Dynamic style injection} gradually refines self-attention interpolation via a scheduling function, balancing style consistency with content fidelity. Extensive experiments demonstrate that our method achieves the highest dual consistency with the fastest inference, delivering high-quality, style-consistent generations without additional fine-tuning or training. We hope this work inspires further research that improves the controllability and efficiency of scale-wise autoregressive T2I models.

\clearpage
\clearpage
\appendix
\maketitlesupplementary

\section{Evaluation Details}
\label{sec:evaluation}
\subsection{Evaluation protocol}
Following the evaluation protocol established by StyleAligned~\cite{hertz2024stylealignedimagegeneration}, we evaluate both the \emph{object relevancy} and \emph{style consistency} of the generated images. 
We further introduce a unified metric, \emph{dual consistency}, to jointly assess these two aspects.

\paragraph{Object relevancy (\(S_\text{obj}\)).}
To evaluate object relevancy, we compute the CLIP cosine similarity~\cite{radford2021learning} between each generated image and its corresponding object description, measuring how faithfully the image reflects the given object prompt. 
We calculate the CLIP-T score for every image–prompt pair and report the average across all samples.

\paragraph{Style consistency (\(S_\text{sty}\)).}
To assess style alignment among images conditioned on the same identity prompt, we compute the average pairwise DINO similarity, following prior work~\cite{hertz2024style, frenkel2024implicitstylecontentseparationusing}.  
Specifically, we use the CLS-token features from DINO ViT-B/8~\cite{caron2021emerging}, which capture global style properties such as texture, rendering, and background.

\paragraph{Dual consistency (\(S_\text{dual}\)).}
To jointly capture object relevancy and style consistency, we compute the harmonic mean of the two metrics:
\begin{equation}
    S_\text{dual} = \frac{2 S_\text{obj} S_\text{sty}}{S_\text{obj} + S_\text{sty}}.
\end{equation}

\paragraph{Implementation.}
We use the official evaluation data provided by StyleAligned~\cite{hertz2024stylealignedimagegeneration}, with minor extensions to incorporate the dual consistency metric. All models are evaluated under the same hardware setup, using a single NVIDIA A6000 GPU and PyTorch for implementation.

\subsection{Implementation details for comparison models}
All comparison models are run using the \textbf{official hyperparameters provided in their public repositories} without additional tuning.

\paragraph{Image-based style-aligned image generation models.}
\label{app:imagebased}

We compare our method against several reference image-based style-aligned image generation models that leverage Stable Diffusion XL \cite{podell2023sdxlimprovinglatentdiffusion} or Muse \cite{chang2023musetexttoimagegenerationmasked} as their backbone. Specifically, we include the following representative models. For a fair comparison, we use each model's first generated output for a given prompt as its reference image, ensuring that all methods operate under the same reference–generation setting without manual selection bias. Specifically, we include the following models: 

\begin{itemize}
    \item \textbf{IP-Adapter}~\cite{ye2023ipadaptertextcompatibleimage}: 
    We use the official code available at \url{https://github.com/tencent-ailab/IP-Adapter}. (SDXL)

    \item \textbf{B-LoRA}~\cite{frenkel2024implicitstylecontentseparationusing}: 
    We use the official code available at \url{https://github.com/yardenfren1996/B-LoRA}. (SDXL)

    \item \textbf{StyleDrop}~\cite{sohn2023styledrop}: 
    We use the unofficial code available at \url{https://github.com/zideliu/StyleDrop-PyTorch}. (MUSE)

    \item \textbf{DB-LoRA}~\cite{ryu2023low}: 
    We use the official code available at \url{https://github.com/huggingface/diffusers/tree/main/examples/dreambooth}. (SDXL)

    \item \textbf{CSGO}~\cite{xing2024csgo}: We use the official code available at \url{https://github.com/instantX-research/CSGO}. (SDXL)
\end{itemize}

\paragraph{Text-based style-aligned image generation models.}
\label{app:others}
We also compare our method against text-based style-aligned image generation models that leverage Stable Diffusion XL \cite{podell2023sdxlimprovinglatentdiffusion} or FLUX \cite{zhang2025alignedgen} as their backbone. Specifically, we include the following models:

\begin{itemize}
    \item \textbf{StyleAligned}~\cite{hertz2024stylealignedimagegeneration}: We use the official code available at \url{https://github.com/google/style-aligned}. (SDXL)
    \item \textbf{AlignedGen}~\cite{zhang2025alignedgen}: We use the official code available at \url{https://github.com/Jiexuanz/AlignedGen}. (FLUX)
\end{itemize}

\section{Comprehensive Analysis of our method}

\subsection{Additional analysis on generation process} To verify whether our method operates as intended during the generation process, we conduct an additional analysis at each step using 400 images generated from our method and the baseline model, Infinity \cite{han2024infinity}. Specifically, we compare the first image in each batch with the subsequent images at different generation steps to examine how RGB statistics, style, and content evolve over time. 

\begin{figure}[t]
    \centering
    \includegraphics[width=1 \linewidth]{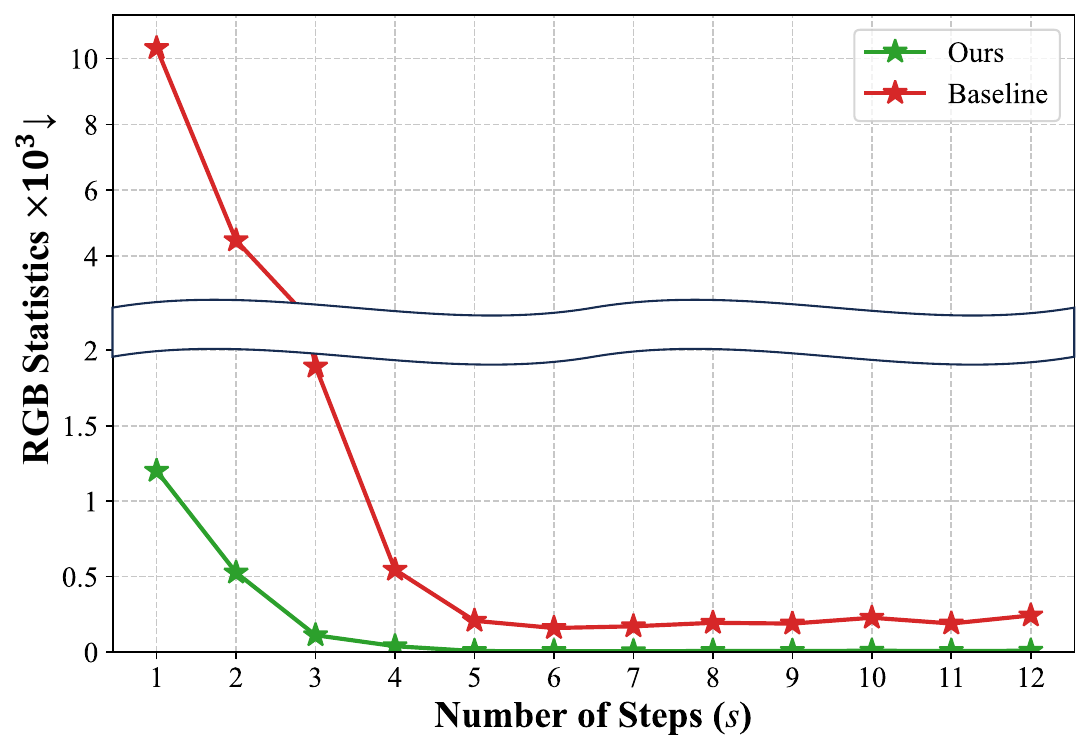}
    \captionof{figure}{Step-wise comparison of RGB statistic changes between our method and the baseline model (Lower is better).
    }
    \vspace{-3mm}
    \label{fig:abl_rgb_step}
\end{figure}

\figref{fig:abl_rgb_step} illustrates the evolution of RGB statistics across generation steps for both our proposed method and the baseline model. Compared to the baseline, our method starts with relatively lower RGB statistics thanks to our initial feature alignment. Furthermore, our approach ensures rapid convergence, with RGB statistics nearly identical after replacement, whereas the baseline fails to fully align them in the later steps. This highlights the effectiveness of our method in enforcing consistency across generated images.

\begin{figure}[t]
    \centering
    \includegraphics[width=1 \linewidth]{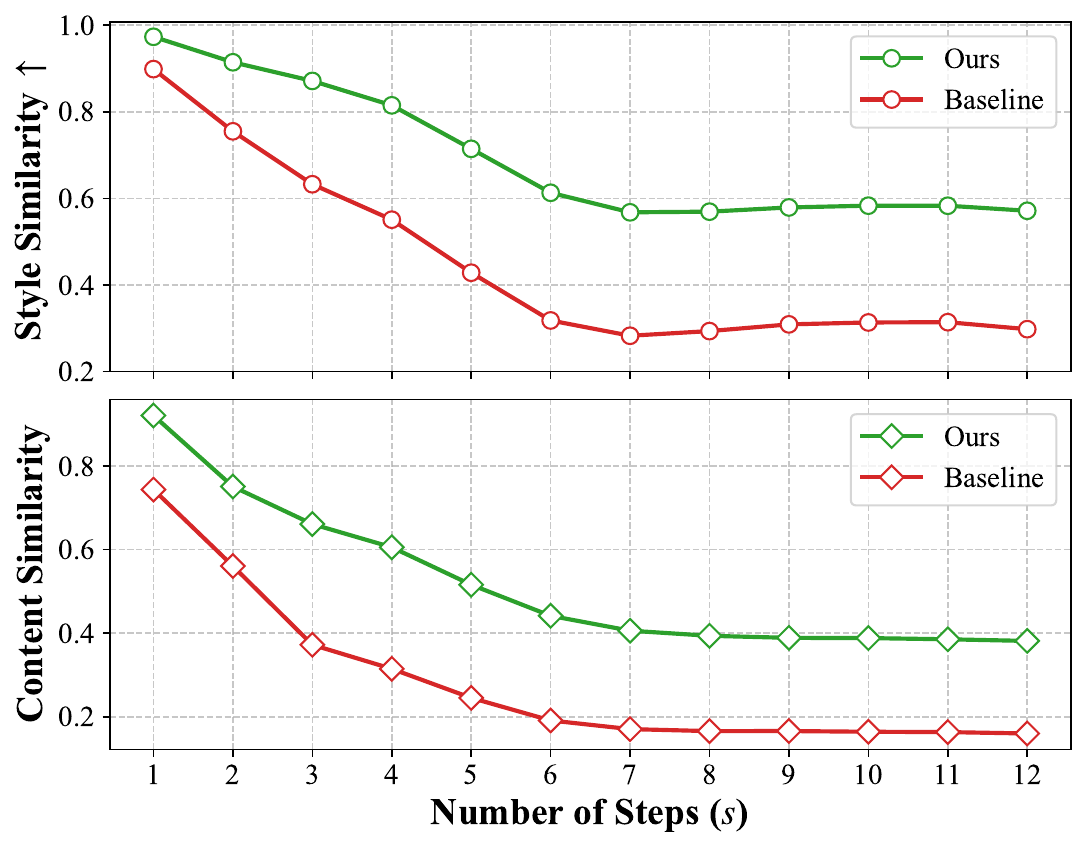}
    \captionof{figure}{Step-wise comparison of content and style similarity between our method and the baseline model (Higher is better).
    }
    
    \label{fig:abl_cnt_sty_step}
\end{figure}

\figref{fig:abl_cnt_sty_step} shows that the baseline model exhibits a sharper decline, indicating significant deviations in content structure and style as generation progresses. In contrast, our method moderates this effect through pivotal feature interpolation and dynamic style injection. While the difference is not overly pronounced, it suggests that our approach maintains a balance—preserving structural coherence while allowing for natural refinement without overly constraining content variation.
Additionally, the improved style similarity demonstrates that our method effectively enforces style consistency across steps without compromising content integrity.

\begin{figure*}[t]
    \centering
    \includegraphics[width=1 \linewidth]{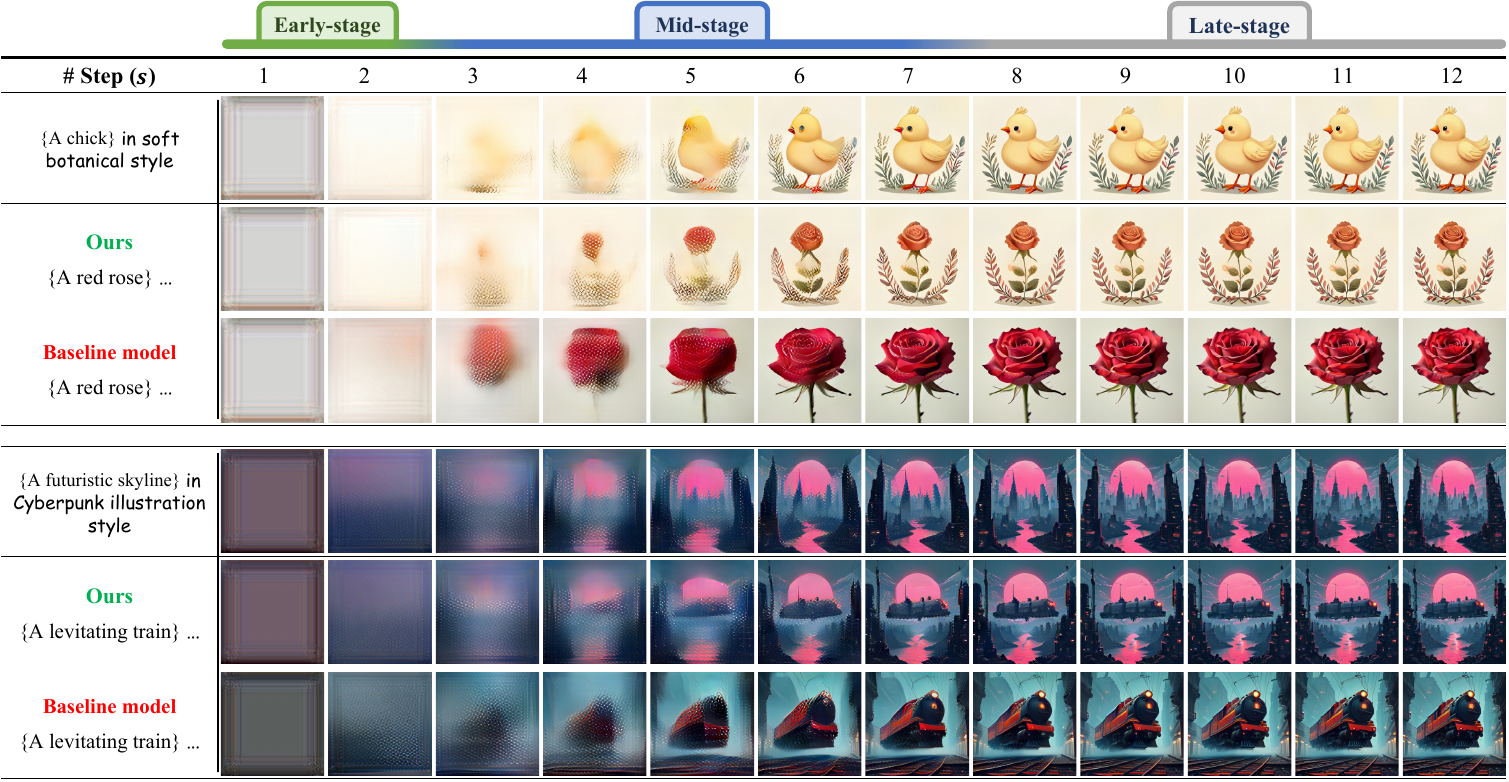}
    \captionof{figure}{Qualitative results of step-wise changes between Standard Text-to-Image model and Ours
    }
    \label{fig:rgb_step_qual}
\end{figure*}

These tendencies are further supported by the qualitative results in \figref{fig:rgb_step_qual}. In the standard text-to-image model, the generated images fail to align with the first-row images in style, resulting in noticeable background color discrepancies. Additionally, key elements such as the ``red rose'' and ``levitating train'' do not adhere to the color scheme or composition of the first-row images, leading to higher RGB statistical deviations. In contrast, our model effectively integrates background colors throughout the generation process while ensuring that key objects maintain consistent positioning and composition, preserving their independence.


\subsection{Analysis on mid-stage pivotal feature interpolation}
\begin{figure}[t]
    \centering
    \includegraphics[width=1 \linewidth]{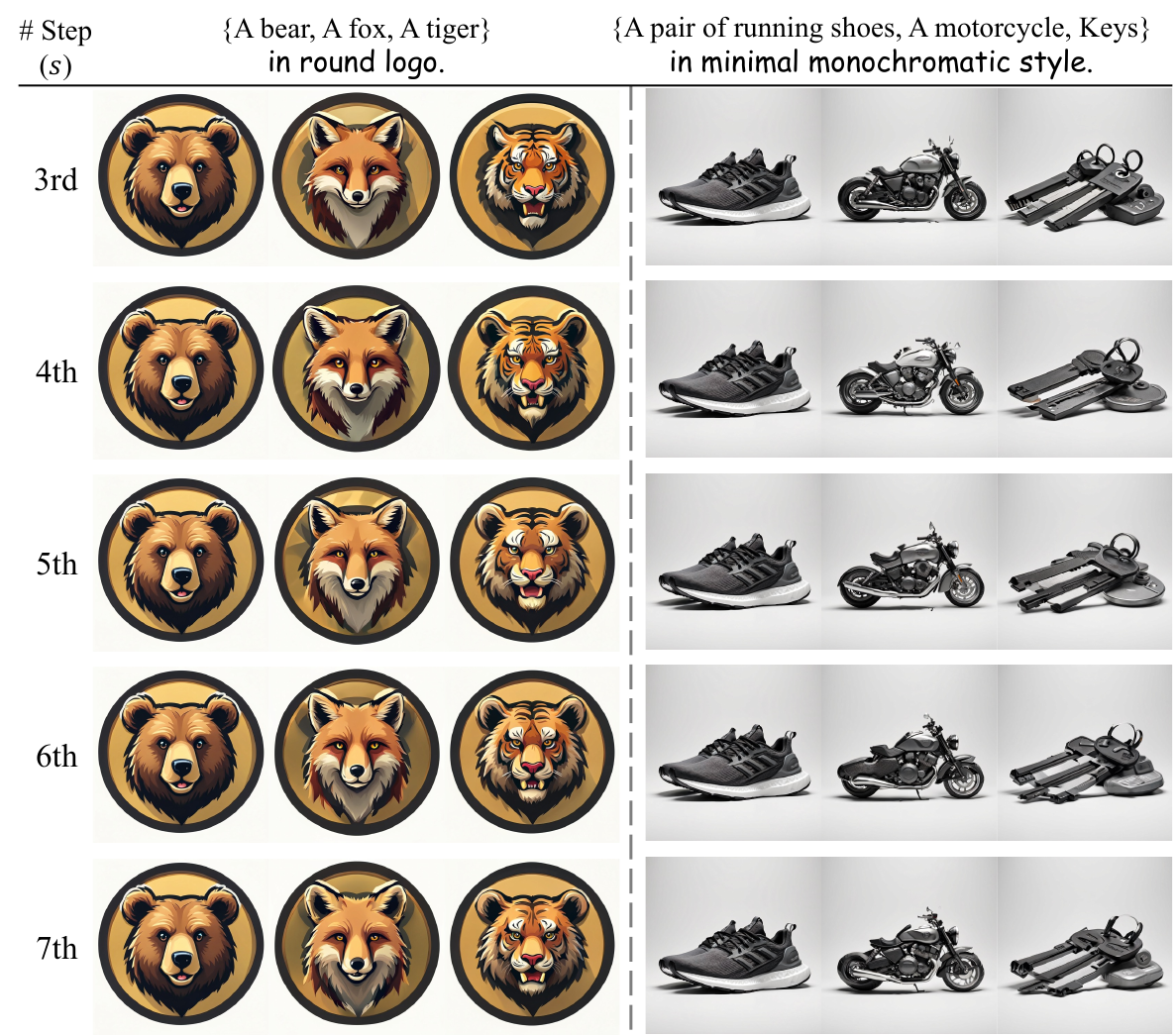}
    \captionof{figure}{Qualitative results analyzing the impact of applying pivotal feature interpolation at different mid-stage steps (e.g., 3rd to 7th). 
    }
    \label{fig:key_feature}
\end{figure}

We conduct additional experiments to assess the impact of applying \textit{pivotal feature interpolation} at different mid-stage steps (\textit{e.g.}, 3rd to 7th). As illustrated in \figref{fig:key_feature}, the results support our hypothesis that the next-scale prediction process plays a crucial role in determining object placement and overall style. Moreover, the results show that pivotal feature interpolation can be applied at any mid-stage step with minimal variation in performance, making it a flexible hyperparameter. This demonstrates the robustness of our model in maintaining style consistency and object alignment across various interpolation points.

\begin{table}[t]
\centering
    \resizebox{0.7\linewidth}{!}{
    \begin{tabular}{cccc}
    \toprule
    Method  & \( S_{\text{dual}}\) $\uparrow$ & \(S_{\text{obj}}\) $\uparrow$ & \(S_{\text{sty}}\) $\uparrow$ \\
    \midrule
    $r$=1 & 0.364 & 0.229 & \textbf{0.891} \\
    $r$=2 & 0.372 & 0.247 & 0.756 \\
    $r$=3 & 0.373 & 0.276 & 0.577 \\
    $r$=3.4$^*$ & \textbf{0.375} & 0.282 & 0.556 \\
    $r$=4 & 0.366 & 0.287 & 0.504 \\
    $r$=5 & 0.356 & \textbf{0.291} & 0.458 \\
    \midrule
    $\alpha_\text{pivot}$=0.2 & 0.370 & \textbf{0.284} & 0.532 \\
    $\alpha_\text{pivot}$=0.3 & 0.371 & 0.283 & 0.536 \\
    $\alpha_\text{pivot}$=0.4$^*$ & \textbf{0.375} & 0.282 & 0.556 \\
    $\alpha_\text{pivot}$=0.5 & 0.372 & 0.277 & 0.567 \\
    $\alpha_\text{pivot}$=0.6 & 0.373 & 0.276 & \textbf{0.601} \\
    \bottomrule
    \end{tabular}
    }
    \caption{Additional ablation study on hyperparameter decay rate $r$ and pivot weight $\alpha_\text{pivot}$. Ours$^*$.}
    \vspace{-12pt}
    \hspace{10pt}
    \label{tab:params_ablation}
\end{table}

\subsection{Hyperparameter robustness}
We also performed ablations on the decay rate $r$ and the pivot weight \(\alpha_{\text{pivot}}\) in \tabref{tab:params_ablation}. The results demonstrate that our method is robust to variations in both $r$ and $\alpha_{\text{pivot}}$ in terms of dual consistency $S_{\text{dual}}$. Additionally, we observe a trade-off between object relevancy and style consistency, highlighting the importance of balancing these aspects.
Thus, we highlight that our method offers flexible control—appropriate hyperparameter tuning enables high object relevancy while preserving strong dual consistency.

\subsection{Quantitative analysis of our observation}
To further support our design choices in \secref{sec:observation} of the main paper, we present a quantitative analysis of the scale-wise generation process, focusing on feature dynamics across early, mid, and late stages.

Specifically, we measure the average differences between consecutive generation steps in both content and style features. The content features are extracted using VGG19 activations, while the style features are obtained via DINO ViT-B/8 embeddings. For each region of the generation process—early (steps 1–2), mid (steps 3–7), and late (steps 8–12)—we compute the mean feature changes as follows:

\begin{itemize}
\item \textbf{Content feature differences}: (Early: 0.031, Mid: 0.137, Late: 0.042)
\item \textbf{Style feature differences}: (Early: 0.002, Mid: 0.158, Late: 0.018)
\end{itemize}

These statistics indicate that both content and style features undergo significantly greater variation during the mid-stage compared to the early or late stages. This validates our hypothesis that the mid-stage plays a pivotal role in defining the spatial layout and visual style of the generated images.

Building on this observation, we propose \textit{pivotal feature interpolation} and \textit{dynamic style injection} specifically at the mid-stage to reinforce style alignment and ensure coherent object placement across the generated image set. The effectiveness of this strategy is further confirmed by both quantitative and qualitative results presented in the main paper and this appendix.

\subsection{Limitation of our method}
\begin{figure}[t]
    \centering
    \includegraphics[width=1 \linewidth]{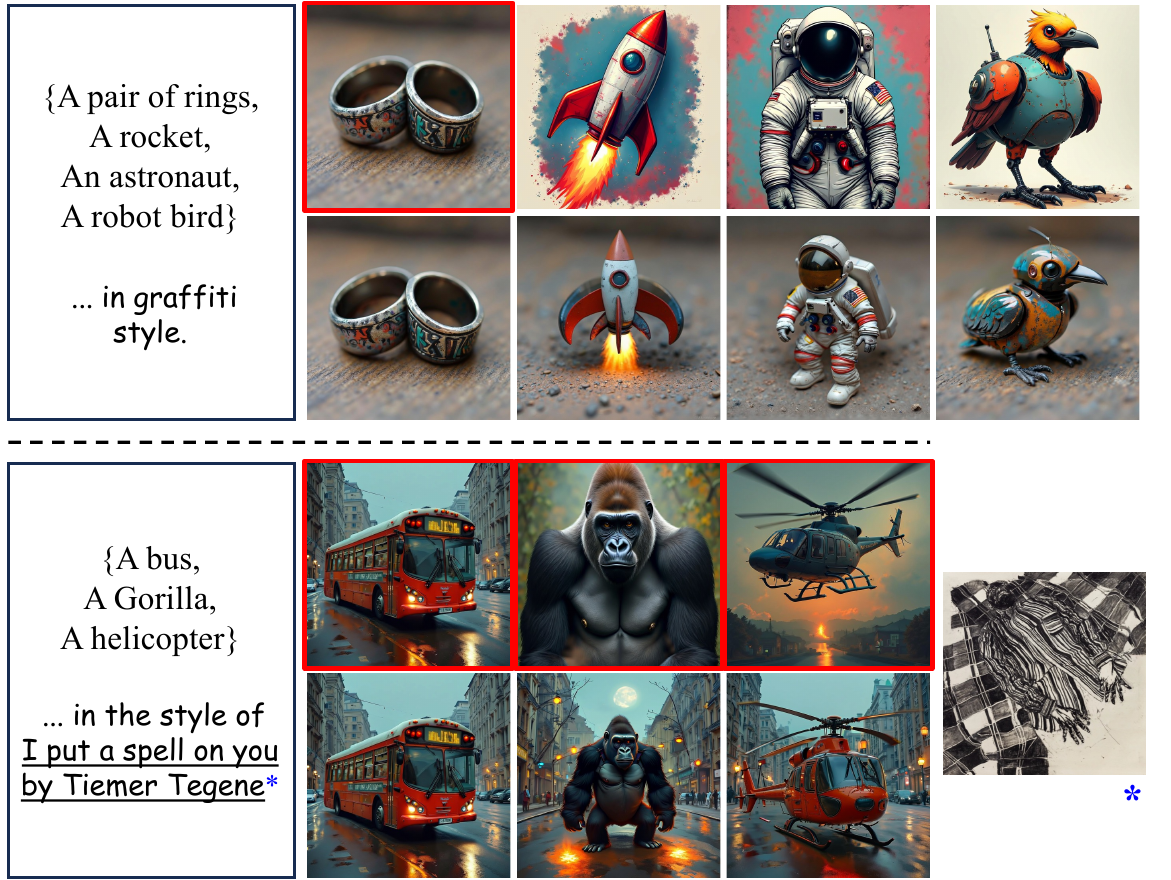}
    \captionof{figure}{Visualization of a limitation in our approach. While our method effectively aligns styles across generated images, it relies on the baseline model’s ability to generate the intended style.
    }
    \label{fig:limitation}
\end{figure}

\figref{fig:limitation} highlights a key limitation of our approach, emphasizing its reliance on the baseline model’s ability to generate the desired style. While our method effectively enforces style consistency across the generated images, it does not modify the style generation process itself. As shown in the first example, when the baseline model fails to correctly apply the specified style to the first image in the batch, our method ensures consistency but propagates the incorrect style to the remaining images. Similarly, in the second example, when the baseline struggles to reproduce the style of a recent artwork, our approach maintains style alignment across images, but the overall style still deviates from the intended prompt. These cases demonstrate that while our method successfully aligns styles within a batch, it cannot correct or enhance the baseline model’s style generation.

\begin{table}[t]
\centering
\resizebox{1\linewidth}{!}{%
\footnotesize
\begin{tabular}
{c|c|ccc}
\toprule
Metric  & \multicolumn{1}{c}{\textbf{Ours}} & \multicolumn{1}{|c}{FireFlow} & \multicolumn{1}{c}{RF-Solver-Edit} & \multicolumn{1}{c}{RF-Inversion} \\
\midrule
    \(S_{\text{obj}}\) $\uparrow$ & \underline{0.282} & 0.281 & \textbf{0.302} & {0.267} \\
    \(S_{\text{sty}}\) $\uparrow$ &  \textbf{0.556} & \underline{0.530} & 0.292 & {0.425} \\
    \(S_{\text{dual}}\) $\uparrow$ &  \textbf{0.375} & \underline{0.367} & 0.270 & {0.328} \\
\midrule
   Time (s) $\downarrow$ &  \textbf{1.58} & \underline{11.25} & 633.20 & 544.06  \\ 
\bottomrule
\end{tabular}
}
\caption{Comparison with recent image editing models.}
\label{tab:add_comp}
\end{table}


\section{Comparison to recent image editing models}
We further conducted comparisons with recent image editing models, including FireFlow \cite{deng2024fireflow}, RF-Solver-Edit \cite{wang2024taming}, and RF-Inversion \cite{rout2024semantic}, to demonstrate the effectiveness of our model. As shown in \tabref{tab:add_comp}, our model consistently outperforms existing methods. Moreover, unlike exemplar-based methods that require reference images, our text-driven approach achieves strong performance while offering greater flexibility and practicality.

\section{Additional style-aligned image generation results}
\label{sec:add_results}
\begin{figure*}[h]
    \centering
    \includegraphics[width=.9 \linewidth]{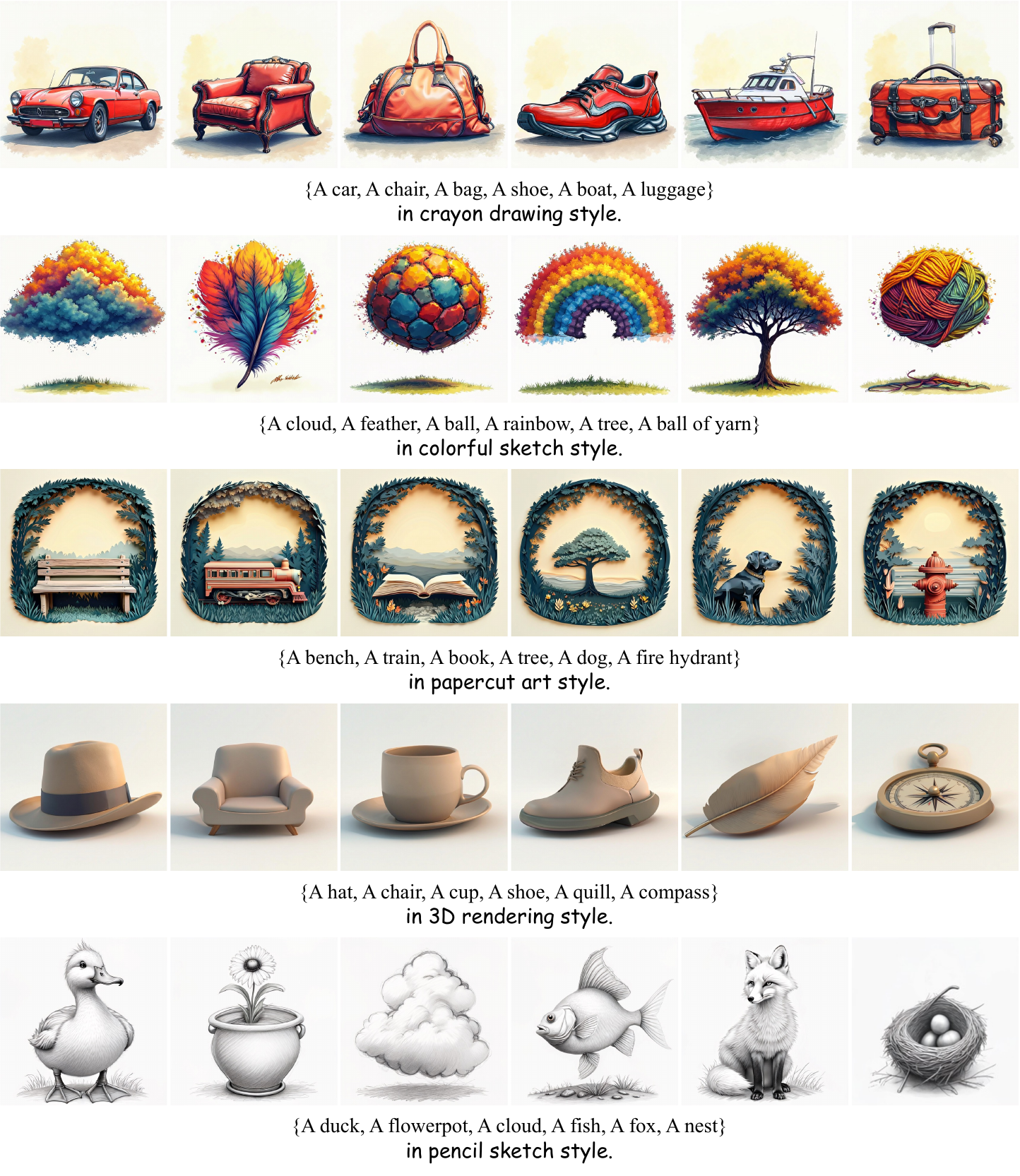}
    \captionof{figure}{Style-aligned image generation results using our model. Each row presents a set of objects rendered in a specific artistic style, including crayon drawings, colorful sketches, papercut art, 3D rendering, and pencil sketches. 
    }
    \label{fig:qual1}
\end{figure*}

\begin{figure*}[h]
    \centering
    \includegraphics[width=.9 \linewidth]{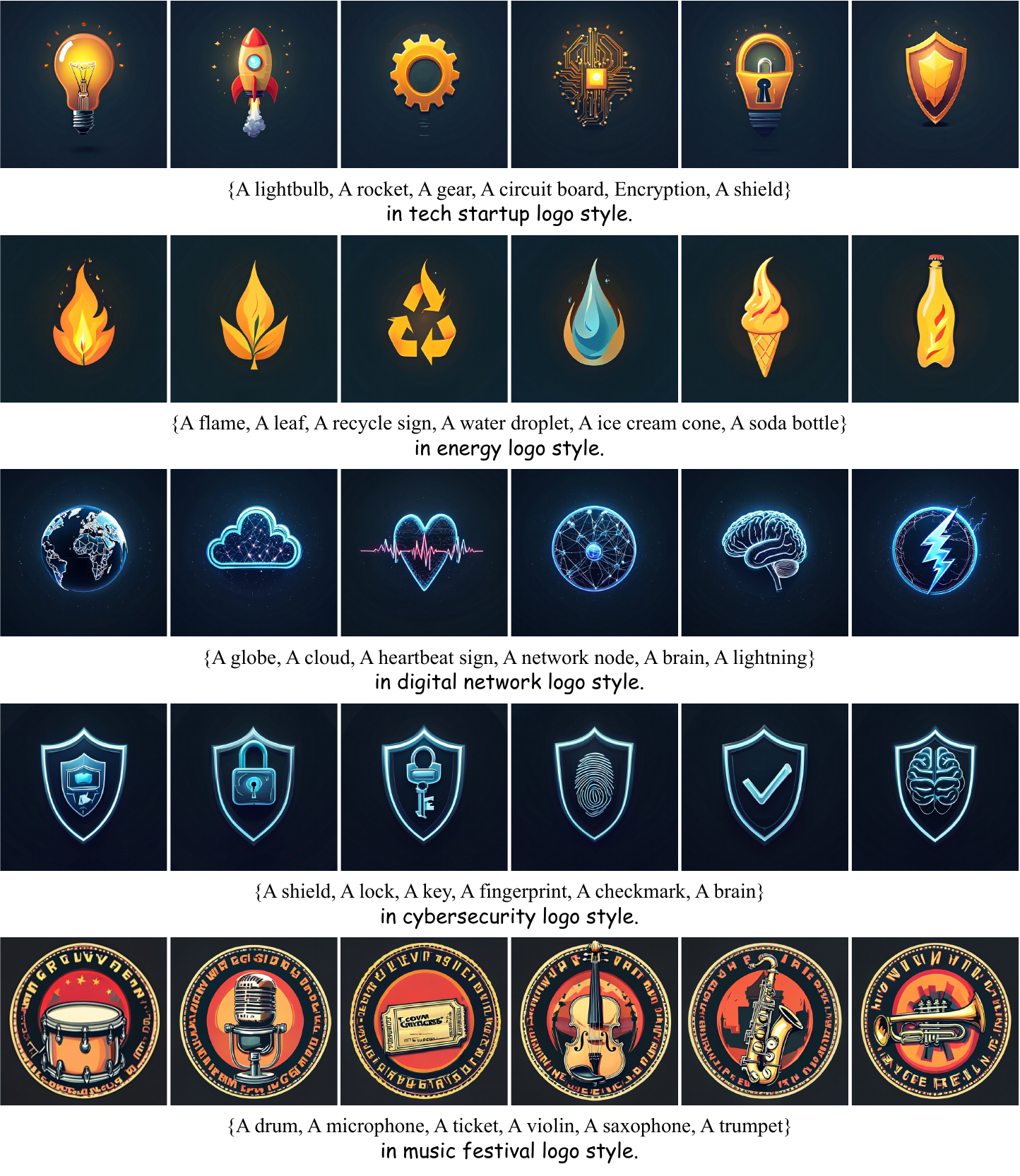}
    \captionof{figure}{Style-aligned image generation results using our model. Each row presents a set of objects rendered in a specific logo style, including tech startup logos, energy logos, digital network logos, cybersecurity logos, and music festival logos styles.
    }
    \label{fig:qual2}
\end{figure*}

\begin{figure*}[h]
    \centering
    \includegraphics[width=.9 \linewidth]{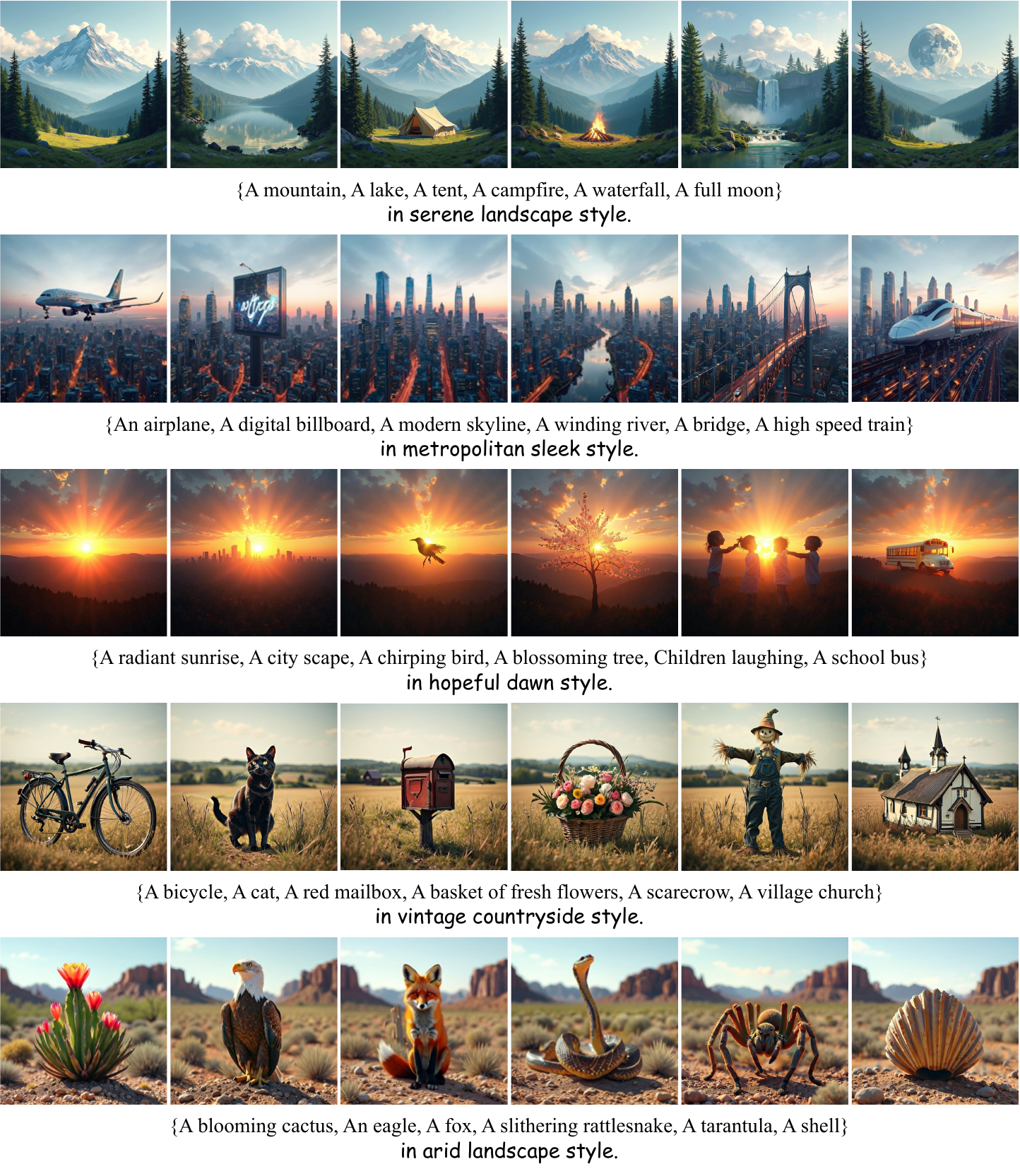}
    \captionof{figure}{Style-aligned image generation results using our model.  Each row showcases a set of objects rendered in a distinct realistic style, including serene landscape, metropolitan sleek, hopeful dawn, vintage countryside, and arid landscape styles.
    }
    \label{fig:qual3}
\end{figure*}

We present additional results of our method in \figref{fig:qual1}, \figref{fig:qual2}, and \figref{fig:qual3}, showcasing its effectiveness across various style-aligned image generation scenarios. These figures highlight the versatility of our approach in maintaining consistent visual attributes while preserving content integrity. Notably, our method demonstrates robustness across different prompts and artistic styles, reinforcing its adaptability to diverse generative tasks.

\clearpage
\clearpage
{
    \small
    \bibliographystyle{ieeenat_fullname}
    \bibliography{main}
}

\end{document}